\documentclass[journal]{IEEEtran}
\IEEEoverridecommandlockouts
\usepackage{cite}
\usepackage{soul}
\usepackage{amsmath,amssymb,amsfonts}
\usepackage{algorithmic}
\usepackage{textcomp}
\usepackage{graphicx}
\usepackage{xcolor}
\usepackage{multirow}
\def\BibTeX{{\rm B\kern-.05em{\sc i\kern-.025em b}\kern-.08em
    T\kern-.1667em\lower.7ex\hbox{E}\kern-.125emX}}

\graphicspath{{figures/}}

\begin{document}

\title{Revisiting the Application of Feature Selection Methods to Speech Imagery BCI Datasets}

\author{Javad Rahimipour Anaraki, Jae Moon, and Tom Chau
\thanks{J. R. A., J. M., and T. C. are with the Institute of Biomedical Engineering, University of Toronto, Rosebrugh Building, 164 College Street, Toronto, Ontario M5S 3G9 Canada and the Bloorview Research Institute, Holland Bloorview Kids Rehabilitation Hospital, 150 Kilgour Road Toronto, Ontario M4G 1R8 Canada (e-mail: j.rahimipour@utoronto.ca; jae.moon@mail.utoronto.ca; and tom.chau@utoronto.ca)}}

\maketitle

\begin{abstract}
Brain-computer interface (BCI) aims to establish and improve human and computer interactions. There has been an increasing interest in designing new hardware devices to facilitate the collection of brain signals through various technologies, such as wet and dry electroencephalogram (EEG) and functional near-infrared spectroscopy (fNIRS) devices. The promising results of machine learning methods have attracted researchers to apply these methods to their data. However, some methods can be overlooked simply due to their inferior performance against a particular dataset. This paper shows how relatively simple yet powerful feature selection/ranking methods can be applied to speech imagery datasets and generate significant results. To do so, we introduce two approaches, horizontal and vertical settings, to use any feature selection and ranking methods to speech imagery BCI datasets. Our primary goal is to improve the resulting classification accuracies from support vector machines, $k$-nearest neighbour, decision tree, linear discriminant analysis and long short-term memory recurrent neural network classifiers. Our experimental results show that using a small subset of channels, we can retain and, in most cases, improve the resulting classification accuracies regardless of the classifier.
\end{abstract}


\section{Introduction}
\IEEEPARstart{B}{rain}-computer interfaces (BCI) enable individuals to control, interact and communicate with their environment through mental activity alone \cite{wolpaw2002brain,mcfarland2011brain}. The first component is a data acquisition method, where brain signals of interest are detected. The second component is the signal processing stage, where brain signals are processed and analyzed before issuing corresponding commands. The final part is the output stage, where detected and processed control signals produce external changes that may be used as feedback to the user \cite{wolpaw2002brain}. Multiple modalities exist for acquiring brain signals of interest, including but not limited to, functional magnetic resonance imaging (fMRI), functional near-infrared spectroscopy (fNIRS), magnetencephalography (MEG), transcranial Doppler (TCD), electrocorticography (ECoG), and electroencephalography (EEG). While each modality has its benefits and caveats, namely portability, cost, ease of setup, invasiveness, and spatial and temporal resolution, EEG is the most popular modality in BCI operation. 

The pre-processing stage is often necessary to extract useful features from acquired brain signals. Such methods may include downsampling, where the frequency at which the data were initially collected is reduced to a lower sampling rate. Besides, filters may be applied to the acquired data to limit their content to a meaningful range. Such methods are implemented to eliminate unwanted noise and to focus on bands of interest. For instance, band-pass filters between 0.5 and 50 Hz have been used for covert speech paradigms that depend on Delta (0-4 Hz), Theta (4-8 Hz), Alpha (8-12 Hz), Beta (12-30 Hz), and low-frequency Gamma-band (30-50 Hz) activities \cite{brigham2010imagined, d2009toward}. Filters can also be implemented for removing artifacts arising from power line noise and eye movements. Signal artifacts due to the latter can also be suppressed via signal decomposition methods such as independent component analysis (ICA) \cite{vigario1997extraction}. In the feature extraction stage of the signal processing pipeline, specific components of the signal are sought to identify particular brain signals that correspond to user intent \cite{wolpaw2002brain, nicolas2012brain}. These features may be time- and/or frequency-dependent. With speech-related signals, features are usually both time and frequency-dependent, as changes in spectral power are dependent on the timing of phonological information \cite{di2015low}.

In machine learning, the process of removing irrelevant and redundant features is called feature selection \cite{hall1998practical}. Feature selection encompasses both the volume and veracity of big data. It reduces the size of the dataset by discarding unnecessary features so that the size of the dataset becomes smaller and chooses important and contributing features that convey meaningful information about the outcome. Feature selection is divided into two main categories, feature subset selection (FSS) and feature ranking (FR). Feature subset selection, which is usually used interchangeably with feature selection, selects a subset of features based on a selection criterion and through a search method. However, feature ranking methods output a ranked list of features, where using a threshold, a subset can be chosen from the list. 

Relief is a well-known and widely used feature ranking method proposed by Kira \cite{kira1992feature} that performs relatively superior in a wide range of datasets \cite{urbanowicz2018relief}. It starts by choosing a random sample from a dataset, with $n$ samples and $f$ features, and calculates its Euclidean distance to two other samples chosen randomly from the positive and negative classes, called \emph{near-hit} and \emph{near-miss}, respectively. Then, it updates the weight vector $W$, and continues for $m$ times. The resulting weight vector $W$ contains values in the range of $[0\: 1]$ for each feature, where those features that have a weight less than a threshold would be discarded as irrelevant features. The complexity of Relief is $O(m \times n \times f)$. 

Minimum redundancy maximum relevance (mRMR) proposed by Peng et al. \cite{peng2005feature} is a feature selection method based on mutual information, aiming to increase selection pressure for relevant features and reduce the presence of redundant features in the final subset. The complexity of mRMR is $O(|S| \times M)$, where $M$ is the number of features in the original dataset.

Laplacian feature selection proposed by He et al. \cite{he2006laplacian} uses Laplacian eigenmaps and locality preserving projection to calculate features' scores of how much locality power they preserve. The process starts by constructing a graph with $s$ nodes, where $s$ is the number of samples in a dataset. There will be a weighted edge between two nodes if either of those is among $k$ nearest neighbour of the other, and the Laplacian score (LS) is calculated based on the resulting weights.

Using recent advancement in brain science and machine learning methods have pushed BCI research remarkably. Many researchers have explored the application of a wide variety of feature selection methods and classifiers to distinguish top contributing signals and improve the resulting classification accuracies \cite{lotte2018review, naseer2015fnirs, ahsan2009emg}. However, in most cases, machine learning methods have been utilized "as is", and not much research has been done in exploiting different aspects of those methods, notably feature selection and ranking and the way they should apply to EEG data.

In this paper, we explore the application of Relief, mRMR and LS to EEG datasets through a simple technique where we apply feature selection to every trial and find the most common features to form a final ranking list. We compare this approach with the one where we merge all the trials and then apply feature selection to the resulting dataset. In the next section, we provide more details on our proposed method and in Section \ref{proposed}. The results and discussion are presented in Section \ref{results}, and we conclude the paper in Section \ref{conclude}.

\section{Proposed Method} \label{proposed}
In a BCI system, the number of trials can vary depending on the task and the approach. Of all types of BCI, speech imagery BCIs have received a great deal of attention due to their potential for re-enabling speech-impaired individuals to communicate with higher degrees of freedom. Since choosing the most contributing channels to the outcome is a significant problem in a BCI system, there have been many attempts where different feature selection methods (not feature ranking) were employed to filter out less informative channels \cite{gonzalez2019new,feng2019optimized,gupta2019utility,jin2019correlation,mishuhina2018feature,acevedo2019comparison}. Relatively, feature ranking methods, such as Chi-squared, mutual information and gain-ratio \cite{guyon2003introduction} has played a smaller role in BCI systems \cite{baig2019filtering}. In the context of speech imagery BCIs, many invoke common spatial patterns (CSP) to select the appropriate channels \cite{dasalla2009single,wu2017spatial}. In this paper, we investigate the applicability of feature ranking and selection methods to speech imagery EEG datasets in BCI systems using a simple frequency analysis of ranking lists resulting from each trial compared to where all the trials are merged to form a single dataset.

\subsection{Horizontal setting}
\begin{figure}[htbp]
\centering
\includegraphics[scale=.85]{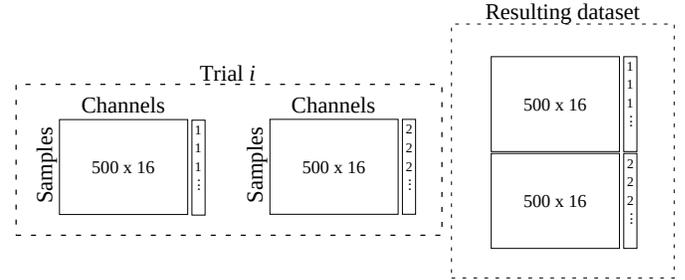}
\caption{Formation of a dataset in horizontal setting for each trial}
\label{horizontal}
\end{figure}

Consider a dataset $D$ with the size of $(s \times c \times t \times c)$, where $s$ is the number of samples, $c$ is the number of channels, $t$ is the number of trials, and $c$ is the number of classes. We want to apply a feature ranking/selection method to each trail to find the ranking list of channels that have better distinguishing power to the others. For each trial, an extra column is added, which contains class labels. To further clarify the setting, consider a dataset with size of $(500 \times 16 \times 5 \times 2)$. Trial $i$ of class one is labelled one and is concatenated with trial $i$ of class two with label two to form a dataset with 1000 samples, 16 channels of trial $i$ containing two classes. Each trial results in a dataset, as shown in Figure \ref{horizontal}.

The resulting dataset is then fed to an arbitrary feature ranking/selection method, and the resulting ranking list of channels is stored in a vector. This process iterates over all the trials, and the resulting vectors are combined vertically to form matrix $R$ with a size of $(16 \times 5)$. For each ranking position (row), we have five values (columns) in $R$; therefore, we need a mechanism to choose one channel as the best one for each position. This is done by finding the most frequent channel in each row of $R$ and storing them in a vector called $F$. Vector $F$ will have one value for each position in the ranking list, where some of the channels could repeat in several positions as they might contribute more to the outcome. To finalize the process, we find the unique values in vector $F$ and return them as the final results. 

\subsection{Vertical setting}
\begin{figure}[htbp]
\centering
\includegraphics[scale=.85]{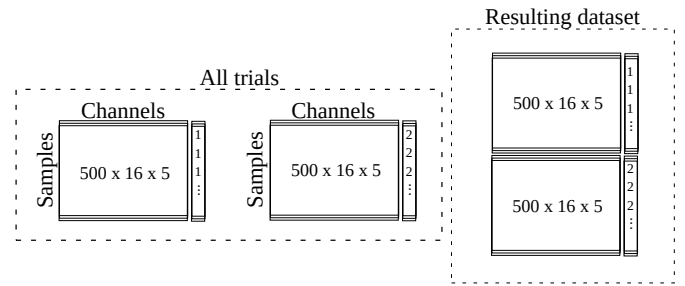}
\caption{Formation of a dataset in horizontal setting for all trials}
\label{vertical}
\end{figure}

In this approach, all the trials for each class are vertically concatenated so that feature ranking/selection can be applied once to the whole dataset (see Figure \ref{vertical}). The advantage of this approach is that all the features are entirely presented to the feature ranking method, which could improve the quality of the selected features in the expense of higher computation time. 

To find the best top $n$ features, we iterate over the ranked list and calculate the resulting classification accuracies. The highest resulting accuracy resulting from each classifier with the corresponding number of selected features is returned as the outcomes.

\section{Experiments}\label{results}

\subsection{Datasets}
To evaluate the performance of the \emph{horizontal setting}, we adopted three datasets (see Table \ref{datasets}) for speech perceptions vs. covert speech called Parallel Linguistic Task (PLT) dataset, KARA ONE and SpeechImagery adopted from PRISM Lab, Computational Linguistics research group\footnote{http://www.cs.toronto.edu/\texttildelow complingweb/data/karaOne/karaOne.html} and Brainliner\footnote{http://brainliner.jp/data/brainliner/speechImageryDataset}, respectively. The goal of the PLT dataset is to distinguish between covert speech and speech presentation of two different colours. KARA ONE \cite{zhao2015classifying} includes data collected to differentiate between vocalized and imagined words. The SpeechImagery dataset \cite{dasalla2009single} consists of the data during vowel speech imagery.

\begin{table}[htbp]
\caption{Datasets' specifications}
\begin{center}
\begin{tabular}{c c c c c}
\hline
\textbf{Dataset} & \textbf{\# Samples} & \textbf{\# Channels} & \textbf{\# Trials} & \textbf{\# Classes}\\\hline
PLT & 2000 & 64 & 100 & 2\\
KARA ONE & 1197 & 64 & 165 & 2\\
SpeechImagery & 128 & 4 & 50 & 2\\\hline
\end{tabular}
\label{datasets}
\end{center}
\end{table}

\subsection{Hardware and software configurations}
We used a Windows 7 machine with Intel\textsuperscript{\textregistered} Core\texttrademark i7-4790S, 16 GB of RAM, using MATLAB\textsuperscript{\textregistered} 9.7.0.1190202 (R2019b) to run all the experiments.

\subsection{Setup}
We split the data to form train and test, each 70\% and 30\%, respectively. All the classification accuracies are computed using support vector machine (SVM) \cite{platt1998fast}, $k$-nearest neighbour ($k$NN)\cite{10.2307/1403796} with $k=3$, decision tree (DT) \cite{quinlan2014c4}, linear discriminant analysis (LDA) \cite{friedman2001elements}, and a long short-term memory (LSTM) recurrent neural network (RNN) classifier using Adam optimization method \cite{kingma2014adam}, 20 epochs, 80 hidden layers and intial learning rate 0.005.

\subsection{Results and Analysis}
To find the best top $n$ features, we iterate over the ranked list generated using Relief, mRMR, and LS and calculate the resulting classification accuracies using SVM, $k$NN, DT, LDA and RNN over all trials of PLT, KARA ONE, and SpeechImagery datasets (see Table \ref{results_horizontal} and \ref{results_vertical} for Horizontal and Vertical settings, respectively).

\begin{table}[htbp]
\caption{Resulting number of selected channels and classification accuracies using SVM, $k$NN and DT for the horizontal setting}
\begin{center}
\setlength{\tabcolsep}{5pt}
\begin{tabular}{c c c c c c}
\hline
\textbf{Dataset} & \textbf{Method} & \textbf{Classifier} & $|\mathbf{\overline{F}}|^a$ & \textbf{CA$^b$} & $\mathbf{\rho}^c$\\\hline
\multirow{15}{*}{PLT} & \multirow{5}{*}{Relief} & SVM & 33.96 & 59.62 (59.32) & 1.76\\
      &  & $k$NN & 27.38 & 99.98 (99.96) & 3.65\\
      &  & DT & 33.88 & 82.80 (82.47) & 2.44\\
      &  & LDA & 32.28 & 58.81 (58.27) & 1.82\\
      &  & RNN & 38.22 & 85.45 (89.40) & 2.24\\\cline{2-6}
      
      & \multirow{5}{*}{mRMR} & SVM & 37.66 & 59.75 (59.32) & 1.59\\
      &  & $k$NN & 28.90 & 99.97 (99.96) & 3.46\\
      &  & DT & 39.90 & 83.14 (82.47) & 2.08\\
      &  & LDA & 37.80 & 59.01 (58.27) & 1.56\\
      &  & RNN & 43.40 & 86.73 (89.40) & 2.00\\\cline{2-6}
      
      & \multirow{5}{*}{LS} & SVM & 32.58 & 59.30 (59.32) & 1.82\\
      &  & $k$NN & 26.46 & 99.95 (99.96) & 3.78\\
      &  & DT & 33.26 & 82.59 (82.47) & 2.48\\
      &  & LDA & 31.62 & 58.57 (58.27) & 1.85\\
      &  & RNN & 37.82 & 83.74 (89.40) & 2.21\\\hline
      
\multirow{15}{*}{KARA ONE} & \multirow{5}{*}{Relief} & SVM & 17.28 & 51.70 (50.48) & 2.99\\
      &  & $k$NN & 39.95 & 83.28 (84.57) & 2.08\\
      &  & DT & 33.25 & 81.29 (80.37) & 2.44\\
      &  & LDA & 25.52 & 51.60 (48.43) & 2.02\\
      &  & RNN & 36.61 & 71.57 (71.55) & 1.96\\\cline{2-6}
      
      & \multirow{5}{*}{mRMR} & SVM & 17.46 & 51.72 (50.48) & 2.96\\
      &  & $k$NN & 39.93 & 83.35 (84.57) & 2.09\\
      &  & DT & 34.46 & 81.60 (80.37) & 2.37\\
      &  & LDA & 41.18 & 52.60 (48.43) & 1.28\\
      &  & RNN & 38.47 & 72.77 (71.55) & 1.99\\\cline{2-6}
      
      & \multirow{5}{*}{LS} & SVM & 19.25 & 51.54 (50.48) & 2.68\\
      &  & $k$NN & 36.55 & 81.39 (84.57) & 2.23\\
      &  & DT & 33.37 & 80.34 (80.37) & 2.41\\
      &  & LDA & 27.35 & 51.62 (48.43) & 1.89\\
      &  & RNN & 35.15 & 70.26 (71.55) & 2.00\\\hline

\multirow{15}{*}{SpeechImagery} & \multirow{5}{*}{Relief} & SVM & 3.22 & 70.50 (69.29) & 21.89\\
      &  & $k$NN & 3.90 & 87.05 (86.89) & 22.32\\
      &  & DT & 3.70 & 79.00 (78.34) & 21.35\\
      &  & LDA & 3.30 & 69.37 (68.47) & 21.02\\
      &  & RNN & 3.08 & 60.68 (57.11) & 19.70\\\cline{2-6}

      & \multirow{5}{*}{mRMR} & SVM & 3.22 & 70.50 (69.29) & 21.89\\
      &  & $k$NN & 3.90 & 87.05 (86.89) & 22.32\\
      &  & DT & 3.70 & 79.00 (78.34) & 21.35\\
      &  & LDA & 3.30 & 69.37 (68.47) & 21.02\\
      &  & RNN & 3.08 & 60.37 (57.11) & 19.60\\\cline{2-6}
      
      & \multirow{5}{*}{LS} & SVM & 3.22 & 70.50 (69.29) & 21.89\\
      &  & $k$NN & 3.90 & 87.05 (86.89) & 22.32\\
      &  & DT & 3.70 & 79.00 (78.34) & 21.35\\
      &  & LDA & 3.30 & 69.37 (68.47) & 21.02\\
      &  & RNN & 3.08 & 61.26 (57.11) & 20.02\\\hline

\end{tabular}
\label{results_horizontal}
\end{center}
$^a$Average of the number of selected features which result in the best classification accuracies (the smaller the better)\\
$^b$Average of the best resulting classification accuracies (in \%) over all trials followed by overall accuracy in presence of all channels (the higher the better)\\
$^c$A measure to consider the performance of feature selection and classifier collectively (the higher the better)
\end{table}

We have applied the vertical setting to the datasets in Table \ref{datasets}, and the results are shown in Table \ref{results_vertical}. To collectively decide on the performance of feature selection methods in combination with different classifiers, we used a measure called $\rho$ as introduced in \cite{anaraki2016new}. To calculate $\rho$, the resulting classification accuracy is divided by the number of features to reflect the performance. Higher classification accuracy resulting from a smaller subset of features is an ideal scenario. However, choosing only one feature is an exception and would not be considered an acceptable result, for which there is no rationale from an expert's point of view.

\begin{table}[htbp]
\caption{Resulting number of selected channels and classification accuracies using SVM, $k$NN and DT for the vertical setting}
\begin{center}
\setlength{\tabcolsep}{5pt}
\begin{tabular}{c c c c c c c}
\hline
\textbf{Dataset} & \textbf{FR} & \textbf{Classifier} & $|\mathbf{F}|^a$ & \textbf{CA$^b$} & $\mathbf{\rho}^c$\\\hline
\multirow{15}{*}{PLT} & \multirow{5}{*}{Relief} & SVM & 9 & 50.67 (49.66) & 5.63\\
      &            & $k$NN & 61 & 99.82 (99.82) & 1.64\\
      &            & DT & 58 & 69.77 (69.77) & 1.20\\
      &            & LDA & 51 & 51.24 (51.16) & 1.01\\
      &            & RNN & 55 & 57.23 (56.98) & 1.04\\\cline{3-6}
      
      & \multirow{5}{*}{mRMR} & SVM & 14 & 50.84 (49.66) & 3.63\\
      &            & $k$NN & 64 & 99.82 (99.82) & 1.56\\
      &            & DT & 60 & 69.46 (69.77) & 1.16\\
      &            & LDA & 37 & 51.20 (51.16) & 1.38\\
      &            & RNN & 55 & 57.17 (56.98) & 1.04\\\cline{3-6}
      
      & \multirow{5}{*}{LS} & SVM & 18 & 50.68 (49.66) & 2.82\\
      &            & $k$NN & 64 & 99.82 (99.82) & 1.56\\
      &            & DT & 56 &  69.76 (69.77) & 1.25\\
      &            & LDA & 45 & 51.30 (51.16) & 1.14\\
      &            & RNN & 63 & 57.19 (56.98) & 0.91\\\hline
      
\multirow{15}{*}{KARA ONE} & \multirow{5}{*}{Relief} & SVM & 58 & 51.10 (49.32) & 0.88\\
      &            & $k$NN & 45 & 84.58 (79.92) & 1.88\\
      &            & DT & 63 & 77.77 (77.70) & 1.23\\
      &            & LDA & 64 & 50.32 (50.32) & 0.79\\
      &            & RNN & 31 & 65.03 ( - ) & 2.10\\\cline{3-6}
      
      & \multirow{5}{*}{mRMR} & SVM & 42 & 50.78 (49.32) & 1.21\\
      &            & $k$NN & 39 & 84.21 (79.92) & 2.16\\
      &            & DT & 60 & 77.87 (77.70) & 1.30\\
      &            & LDA & 63 & 50.33 (50.32) & 0.80\\
      &            & RNN & 18 & 68.21 ( - ) & 3.79\\\cline{3-6}
      
      & \multirow{5}{*}{LS} & SVM & 51 & 50.97 (49.32) & 1.00\\
      &            & $k$NN & 35 & 82.96 (79.92) & 2.37\\
      &            & DT & 64 & 77.70 (77.70) & 1.21\\
      &            & LDA & 61 & 50.34 (50.32) & 0.83\\
      &            & RNN & 26 & 72.60 ( - ) & 3.79\\\hline
      
\multirow{15}{*}{SpeechImagery} & \multirow{5}{*}{Relief} & SVM & 2 & 53.39 (51.33) & 26.70\\
      &            & $k$NN & 4 & 60.86 (60.86) & 15.22\\
      &            & DT & 4 & 59.61 (59.61) & 14.90\\
      &            & LDA & 2 & 51.98 (50.63) & 25.99\\
      &            & RNN & 1 & 56.88 (55.78) & 56.88\\\cline{3-6}
      
      & \multirow{5}{*}{mRMR} & SVM & 2 & 53.39 (51.33) & 26.70\\
      &            & $k$NN & 4 & 60.86 (60.86) & 15.22\\
      &            & DT & 4 & 59.61 (59.61) & 14.90\\
      &            & LDA & 2 & 51.98 (50.63) & 25.99\\
      &            & RNN & 1 & 56.88 (55.78) & 56.88\\\cline{3-6}
      
      & \multirow{5}{*}{LS} & SVM & 1 & 53.54 (51.33) & 53.54\\
      &            & $k$NN & 4 & 60.86 (60.86) & 15.22\\
      &            & DT & 4 & 59.77 (59.61) & 14.94\\
      &            & LDA & 2 & 51.98 (50.63) & 25.99\\
      &            & RNN & 4 & 55.78 (55.78) & 13.95\\\hline
\end{tabular}
\label{results_vertical}
\end{center}
$^a$The best number of selected features which result in the best classification accuracies (the smaller the better)\\
$^b$The best resulting classification accuracy (in \%) after applying feature ranking followed by overall accuracy in presence of all channels (the higher the better)\\
$^c$A measure to consider the performance of feature selection and classifier collectively (the higher the better)
\end{table}

The combination of any of the three feature selection methods and $k$NN works the best for PLT. If we follow the results of the measure, $\rho$ for KARA ONE dataset, the combination of Relief and mRMR and LS with SVM outperforms the others; however, by roughly doubling the number of selected channels we see a considerable jump, around 30\%, in the resulting classification accuracies using $k$NN. For the SpeechImagery dataset, the combination of all methods with the classifiers returns shows no significant variation due to the size of the dataset, where the best accuracy dominantly achieved by $k$NN. Using LS, an example of an unsupervised feature selection method, generated impressive results as no outcome was provided. The results of LS and the classifiers are significant compared to the ones. 

Based on the resulting accuracies using three classifiers, it turns out that more complex classifiers tend to reach their best performance with a smaller number of channels. In all cases, SVM and RNN have followed the same pattern where their average accuracies among all trials gained with the small subsets of channels for KARA ONE and SpeechImagery datasets, respectively. The quality of collected data can be why very small subsets do not provide enough information to SVM and RNN to reach their best performance. Moreover, it is very likely that due to the nature of the considered datasets, which are relatively homogeneous with negligible variations among trials of the same subject for the same task, the results are not the best accuracies. For the $k$NN and DT classifiers, the results are significantly improved as both are less complex classifiers. Furthermore, $k$NN is a non-parametric classifier that generally performs superior when dealing with small and less complicated and linearly separable datasets , which follow cluster pattern samples.

In the vertical setting, SVM and RNN show significant results using a smaller number of features to achieve higher classification accuracies. For instance, in SpeechImagery datasets, LS and SVM, and Relief and mRMR and RNN have used only one feature. Adding more features to the subset could not improve the result further, particularly in EEG datasets using relatively complex non-linear classifiers. By solely looking at the $\rho$ values, but the SVM and RNN results, $k$NN outperforms others in PLT and KARA ONE datasets, followed by LDA for SpeechImagery.

One by-product of this research could be locating redundant (correlated) features/channels that can be used to decide how well the electrodes are being used, and remove unused and slightly ones. Furthermore, the degree of redundancy may be attributed to the intrinsic similarities between speech imagery and speech perception, as exemplified by the work of Moon et al. [in-progress] and other neurophysiological evidence which also portrays the two tasks as producing the same source locations \cite{okada2006left} and activation patterns \cite{tian2010mental}.

One limitation of this study is the number of datasets as well as their sizes. Generally, highly dimensional datasets would be ideal for determining which combinations of features work the best for each case. Moreover, a comprehensive study should be done in terms of feature selection and classification methods to decide the applicability of a broader range of algorithms to EEG datasets.

\section{Conclusion}\label{conclude}
Feature selection/ranking is the most widely used as a pre-processing method to find the most critical and informative subset of features with the hope to retain and potentially improve the resulting classification accuracy by reducing the effect of noise and minimizing model complexity. BCI datasets have been introduced a very niche area for applying machine learning and pattern recognition methods to clean, extract and classify information sourcing form brain activities. In this paper, we investigated two approaches to use any feature selection/ranking method to BCI datasets steered by two settings. The first setting is to form sub-datasets by concatenating each class's trial to the other ones, perform feature selection/ranking to that subset, store the ranking list, and form a matrix of all ranked features for all the trials. Then, find the most frequently contributed feature for each position and return one final ranked list of features. In the second setting, we merged all the trials and for a single dataset and use feature selection/ranking to find essential features and then fed them to a classifier. The results show that generally, $k$NN works considerably better with the selected feature using Relief, mRMR and LS methods. 

\section{Acknowledgement}
This work was supported partially by by Mitacs through the Mitacs Elevate program and Holland Bloorview Kids Rehabilitation Hospital Foundation.

\bibliographystyle{ieeetr}
\bibliography{mybib}

\end{document}